\documentclass[10pt,twocolumn,letterpaper]{article}

\usepackage{cvpr}
\usepackage{times}
\usepackage{epsfig}
\usepackage{graphicx}
\usepackage{amsmath}
\usepackage{amssymb}
\usepackage{nccmath}

\usepackage[breaklinks=true,bookmarks=false]{hyperref}

\cvprfinalcopy 
\def\cvprPaperID{2134} 

\ifcvprfinal\fi
\begin{document}

\title{Attention-Based Multimodal Fusion for Video Description}
\author{Chiori Hori
\and Takaaki Hori
\and Teng-Yok Lee
\and Kazuhiro Sumi\thanks{On sabbatical from Aoyama Gakuin University, sumi@it.aoyama.ac.jp}
\and John R. Hershey
\and Tim K. Marks\\
Mitsubishi Electric Research Laboratories (MERL)\\
{\tt\small \{chori, thori, tlee, sumi, hershey, tmarks\}@merl.com}
}

\maketitle

\begin{abstract}
Currently successful methods for video description are based on encoder-decoder sentence generation using recurrent neural networks (RNNs). Recent work has shown the advantage of integrating temporal and/or spatial attention mechanisms into these models, in which the decoder network predicts each word in the description by selectively giving more weight to encoded features from specific time frames (temporal attention) or to features from specific spatial regions (spatial attention). 
In this paper, we propose to expand the attention model to selectively attend not just to specific times or spatial regions, but to specific modalities of input such as image features, motion features, and audio features. Our new modality-dependent attention mechanism, which we call multimodal attention, provides a natural way to fuse multimodal information for video description. 
We evaluate our method on the Youtube2Text dataset, achieving results that are competitive with current state of the art. More importantly, we demonstrate that our model incorporating multimodal attention as well as temporal attention significantly outperforms the model that uses temporal attention alone.
\end{abstract}

\section{Introduction and Related Work}
Automatic video description, also known as video captioning, refers to the automatic generation of a natural language description (e.g., a sentence) that summarizes an input video. Video description has widespread applications including video retrieval, automatic description of home movies or online uploaded video clips, and video descriptions for the visually impaired. Moreover, developing systems that can describe videos may help us to elucidate some key components of general machine intelligence. Video description research depends on the availability of videos labeled with descriptive text. A large amount of such data is becoming available in the form of audio description prepared for visually impaired users.  Thus there is an opportunity to make significant progress in this area.  We propose a video description method that uses an attention-based encoder-decoder network to generate sentences from input video. 

Sentence generation using an encoder-decoder architecture was originally used for neural machine translation (NMT), in which sentences in a source language are converted into sentences in a target language \cite{DBLP:conf/cvpr/VinyalsTBE15, DBLP:conf/emnlp/ChoMGBBSB14}. In this paradigm, the encoder takes an input sentence in the source language and maps it to a fixed-length feature vector in an embedding space. The decoder uses this feature vector as input to generate a sentence in the target language. However, the fixed length of the feature vector limited performance, particularly on long input sentences, so~\cite{NN_MT@ICLR2015} proposed to encode the input sentence as a sequence of feature vectors, employing a recurrent neural network (RNN)-based soft attention model to enable the decoder to pay attention to features derived from specific words of the input sentence when generating each output word.

The encoder-decoder based sequence to sequence framework has been applied not only to machine translation but also to other application areas including speech recognition~\cite{attention4ASR}, image captioning~\cite{DBLP:conf/cvpr/VinyalsTBE15}, and dialog management~\cite{endecoder4dl@sigdial}.

In image captioning, the input is a single image, and the output is a natural-language description. Recent work on RNN-based image captioning includes \cite{DBLP:journals/corr/MaoXYWY14a, DBLP:conf/cvpr/VinyalsTBE15}. To improve performance,~\cite{attention4ic@ICML2015} added an attention mechanism, to enable focusing on specific parts of the image when generating each word of the description.

Encoder-decoder networks have also been applied to the task of video description~\cite{DBLP:conf/naacl/VenugopalanXDRM15}.
In this task, the inputs to the encoder network are video information features that may include static image features extracted using convolutional neural networks (CNNs), temporal dynamics of videos extracted using spatiotemporal 3D CNNs~\cite{C3D}, dense trajectories~\cite{dense_trajectory}, optical flow, and audio features \cite{audio4vd}. 
The decoder network takes the encoder outputs and generates word sequences based on language models using recurrent neural networks (RNNs) based on long short-term memory (LSTM) units~\cite{LSTM} or gated recurrent units (GRUs)~\cite{GRU}. Such systems can be trained end-to-end using videos labeled with text descriptions.

One inherent problem in video description is that the sequence of video features and the sequence of words in the description are not synchronized. In fact, objects and actions may appear in the video in a different order than they appear in the sentence. When choosing the right words to describe something, only the features that directly correspond to that object or action are relevant, and the other features are a source of clutter. It may be possible for an LSTM to learn to selectively encode different objects into its latent features and remember them until they are retrieved.  However, attention mechanisms have been used to boost the network's ability to retrieve the relevant features from the corresponding parts of the input, in applications such as machine translation~\cite{NN_MT@ICLR2015}, speech recognition~\cite{attention4ASR}, image captioning~\cite{attention4ic@ICML2015}, and dialog management~\cite{attention4dm@SLT2016}.   In recent work, these attention mechanisms have been applied to video description \cite{vd2montreal@ICCV2015,attention4spatial@CVPR2016}. Whereas in image captioning the attention is spatial (attending to specific regions of the image), in video description the attention may be temporal (attending to specific time frames of the video) in addition to (or instead of) spatial.   

In this work, we propose a new use of attention: to fuse information across different modalities.  Here we use modality loosely to refer to different types of features derived from the video, such as appearance, motion, or depth,  as well as features from different sensors such as video and audio features.  
Video descriptions can include a variety of descriptive styles, including abstract descriptions of the scene, descriptions focused on objects and their relations, and descriptions of action and motion, including both motion in the scene and camera motion.   The soundtrack also contains audio events that provide additional information about the described scene and its context.   
Depending on what is being described, different modalities of input may be important for selecting appropriate words in the description. For example, the description ``A boy is standing on a hill" refers to objects and their relations. In contrast, ``A boy is jumping on a hill" may rely on motion features to determine the action. "A boy is listening to airplanes flying overhead" may require audio features to recognize the airplanes, if they do not appear in the video.  Not only do the relevant modalities change from sentence to sentence, but also from word to word, as we move from action words that describe motion to nouns that define object types. Attention to the appropriate modalities, as a function of the context, may help with choosing the right words for the video description.

Often features from different modalities can be complementary, in that either can provide reliable cues at different times for some aspect of a scene. Multimodal fusion is thus an important longstanding strategy for robustness.  
However, optimally combining information requires estimating the reliability of each modality, which remains a challenging problem. 
In this work, we propose that this estimation be performed by the neural network, by means of an attention mechanism that operating across different modalities (in addition to any spatio-temporal attention).   By training the system end-to-end to perform the desired description of the semantic content of the video, the system can learn to use attention to fuse the modalities in a context-sensitive way.   We present experiments showing that incorporating multimodal attention, in addition to temporal attention, significantly outperforms a corresponding model that uses temporal attention alone.

\section{Encoder-decoder-based sentence generator}
\label{sec:encdec}
\begin{figure}[t]
	\centering
	\centerline{\includegraphics[width=8.5cm]{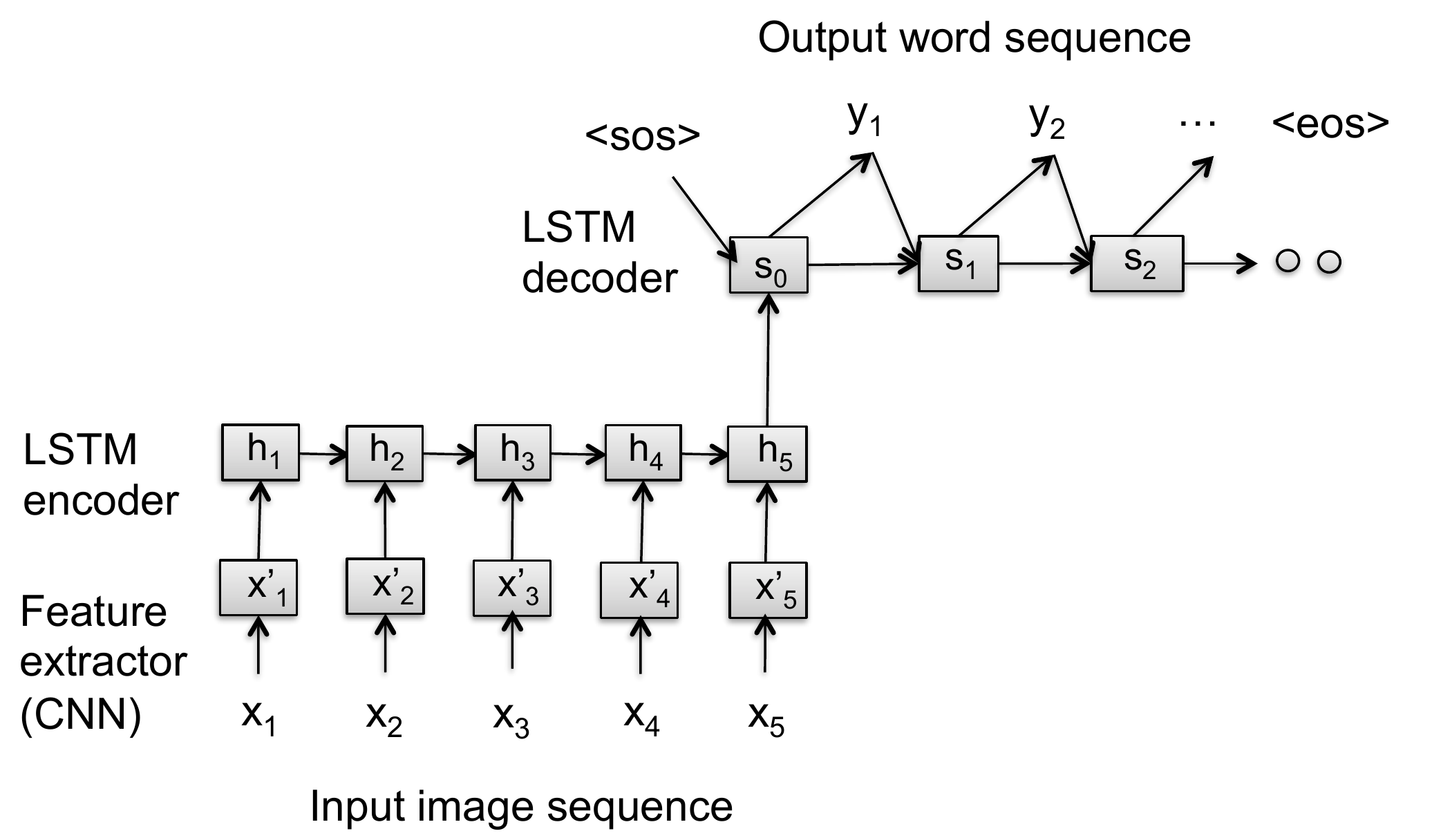}}
	\caption{An encoder-decoder based video description generator.}
	\label{fig:encoder-decoder}
\end{figure}

One basic approach to video description is based on sequence-to-sequence learning. The input sequence, i.e., image sequence, is first encoded to a fixed-dimensional semantic vector. Then the output sequence, i.e., word sequence, is generated from the semantic vector. In this case, both the encoder and the decoder (or generator) are usually modeled as Long Short-Term Memory (LSTM) networks. Figure~\ref{fig:encoder-decoder} shows an example of the LSTM-based encoder-decoder architecture.

Given a sequence of images, $X=x_1,x_2,\dots,x_L$, each image is first fed to a feature extractor, which can be a pre-trained CNN for an image or video classification task such as GoogLeNet \cite{GoogleNet}, VGGNet \cite{VggNet}, or C3D \cite{C3D}.
The sequence of image features, $X'=x'_1,x'_2,\dots,x'_L$, is obtained by extracting the activation vector of a fully-connected layer of the CNN for each input image.\footnote{In the case of C3D, multiple images are fed to the network at once to capture dynamic features in the video.}
The sequence of feature vectors is then fed to the LSTM encoder, and the hidden state of the LSTM is given by
\begin{equation}
h_t=\text{LSTM}(h_{t-1}, x'_t; {\lambda}_E),
\end{equation}
where the LSTM function of the encoder network ${\lambda}_E$ is computed as
\begin{fleqn}[5pt]
\begin{align}
\text{LSTM}(h_{t-1},&x_t;\lambda) = o_t\tanh(c_t),\\
\text{where} \quad 
o_t & = \sigma \bigl( W_{xo}^{(\lambda)}x_t+W_{ho}^{(\lambda)}h_{t-1}+b_o^{(\lambda)} \bigr ) \\
c_t & = f_t c_{t-1} +i_t\tanh \big (W_{xc}^{(\lambda)}x_t \nonumber \\
    &  \hspace{2cm} +W_{hc}^{(\lambda)}h_{t-1}+b_c^{(\lambda)} \big ) \\
f_t & = \sigma \bigl (W_{xf}^{(\lambda)}x_t+W_{hf}^{(\lambda)}h_{t-1}+b_f^{(\lambda)} \bigr ) \\
i_t & = \sigma \bigl (W_{xi}^{(\lambda)}x_t+W_{hi}^{(\lambda)}h_{t-1}+b_i^{(\lambda)} \bigr ), \label{eqn:lstm-end}
\end{align}
\end{fleqn}
where $\sigma()$ is the element-wise sigmoid function, and $i_t$, $f_t$, $o_t$ and $c_t$ are, respectively, the input gate, forget gate, output gate, and cell activation vectors for the $t$th input vector.
The weight matrices $W_{zz}^{(\lambda)}$ and the bias vectors $b_z^{(\lambda)}$ are identified by the subscript $z\in \{x, h, i, f, o, c\}$. For example, $W_{hi}$ is the hidden-input gate matrix and $W_{xo}$ is the input-output gate matrix.
We did not use peephole connections in this work.

The decoder predicts the next word iteratively beginning with the start-of-sentence token, ``\verb+<sos>+'' until it predicts the end-of-sentence token, ``\verb+<eos>+.''
Given decoder state $s_{i-1}$, the decoder network ${\lambda}_D$ infers the next word probability distribution as
\begin{equation}
P(y|s_{i-1})=\text{softmax}\left(W_s^{({\lambda}_D)}s_{i-1} + b_s^{({\lambda}_D)}\right),
\label{eqn:output-dist}
\end{equation}
and generates word $y_i$, which has the highest probability, according to
\begin{equation}
y_i=\mathop{\text{argmax}}_{y \in V} P(y|s_{i-1}),
\label{eqn:output-label}
\end{equation}
where $V$ denotes the vocabulary.
The decoder state is updated using the LSTM network of the decoder as
\begin{equation}
s_i=\text{LSTM}(s_{i-1}, y'_i; {\lambda}_D),
\label{eqn:decoder-state}
\end{equation}
where $y'_i$ is a word-embedding vector of $y_m$,
and the initial state $s_0$ is obtained from the final encoder state $h_L$ and $y'_0=\text{Embed}(\verb+<sos>+)$ as in Figure~\ref{fig:encoder-decoder}.

In the training phase, $Y=y_1,\dots,y_M$ is given as the reference.
However, in the test phase, the best word sequence needs to be found based on
\begin{align}
\hat{Y}&=\mathop{\text{argmax}}_{Y\in V^*} P(Y|X) \\
       &=\mathop{\text{argmax}}_{y_1,\dots,y_M \in V^*} P(y_1|s_0)P(y_2|s_1) \cdots \nonumber \\
       & \hspace{2cm} P(y_M|s_{M-1})P(\mbox{\tt <eos>}|s_M).
\end{align}
Accordingly, we use a beam search in the test phase to keep multiple states and hypotheses with the highest cumulative probabilities at each $m$th step,
and select the best hypothesis from those having reached the end-of-sentence token.

\section{Attention-based sentence generator}
\label{sec:attention}
Another approach to video description is an attention-based sequence generator \cite{NIPS2015_5847}, which enables the network to emphasize features from specific times or spatial regions depending on the current context, enabling the next word to be predicted more accurately.
Compared to the basic approach described in Section~\ref{sec:encdec}, the attention-based generator can exploit input features selectively according to the input and output contexts.
The efficacy of attention models has been shown in many tasks such as machine translation \cite{NN_MT@ICLR2015}.

\begin{figure}[t]
	\centering
	\centerline{\includegraphics[width=8cm]{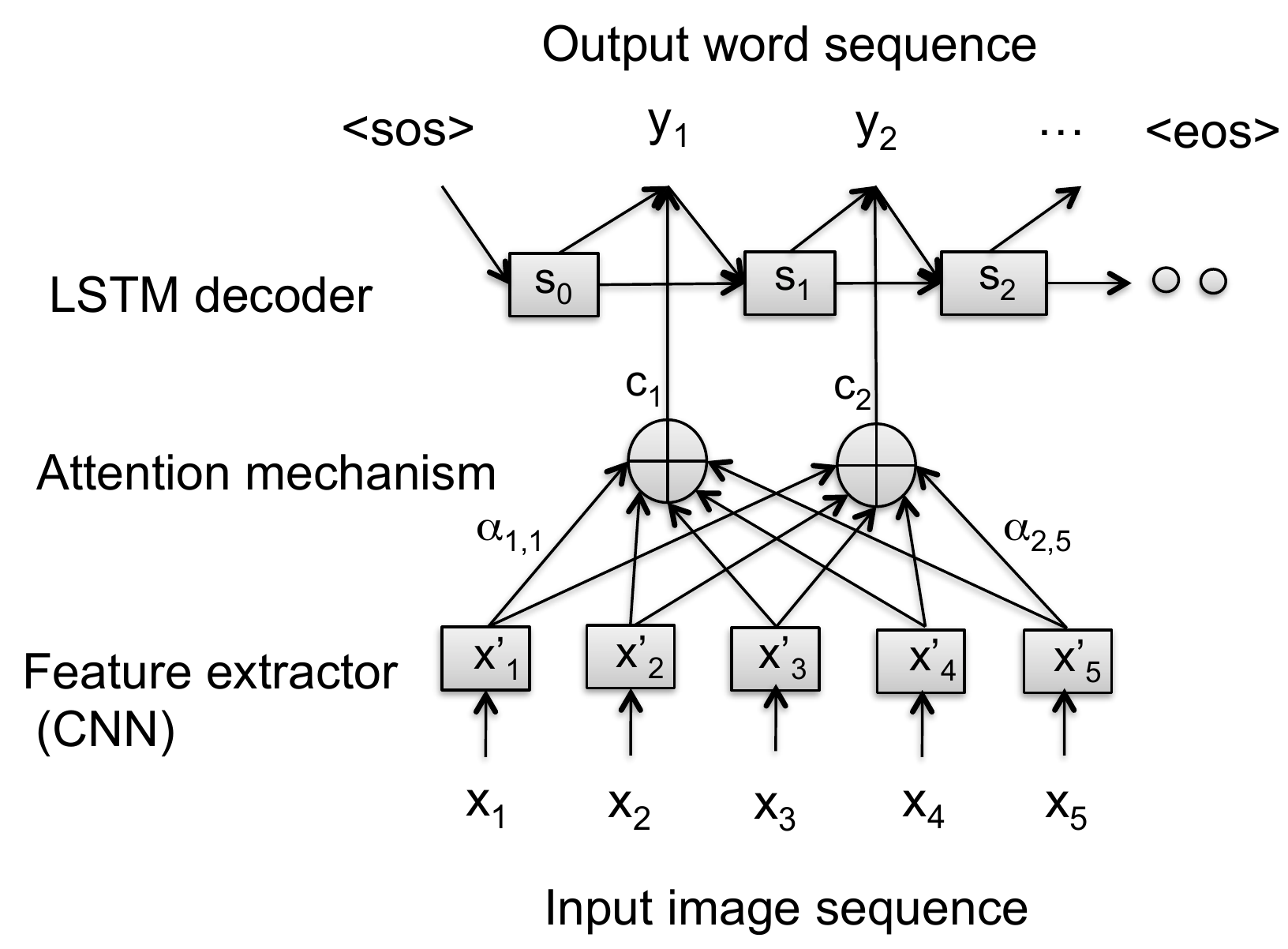}}
	\caption{An encoder-decoder based sentence generator with temporal attention mechanism.}
	\label{fig:attention2}
\end{figure}

Figure~\ref{fig:attention2} shows an example of the attention-based sentence generator from video, which has a temporal attention mechanism over the input image sequence.

The input sequence of feature vectors is obtained using one or more feature extractors.
Generally, attention-based generators employ an encoder based on a bidirectional LSTM (BLSTM) or Gated Recurrent Units (GRU) to further convert the feature vector sequence so that each vector contains its contextual information.
In video description tasks, however, CNN-based features are often used directly, or one more feed-forward layer is added to reduce the dimensionality.

If we use an BLSTM encoder following the feature extraction, then the activation vectors (i.e., encoder states) are obtained as
\begin{equation}
h_t=\left[
\begin{array}{cc}
h^{(f)}_{t}  \\
h^{(b)}_{t} 
\end{array}
\right],
\label{eqn:bi-hidden}
\end{equation}
where $h^{(f)}_{t}$ and $h^{(b)}_{t}$ are the forward and backward hidden activation vectors:
\begin{align}
h^{(f)}_{t}=\mbox{LSTM}(h^{(f)}_{t-1}, x'_t;{\lambda}_E^{(f)}) \\
h^{(b)}_{t}=\mbox{LSTM}(h^{(b)}_{t+1}, x'_t;{\lambda}_E^{(b)}).
\end{align}
If we use a feed-forward layer, then the activation vector is calculated as
\begin{equation}
h_t = \tanh(W_p x'_t + b_p),
\end{equation}
where $W_p$ is a weight matrix and $b_p$ is a bias vector.
If we use the CNN features directly, then we assume $h_t=x'_t$.

The attention mechanism is realized by using {\it attention weights} to the hidden activation vectors throughout the input sequence. These weights enable the network to emphasize features from those time steps that are most important for predicting the next output word.

Let $\alpha_{i,t}$ be an attention weight between the $i$th output word and the $t$th input feature vector.
For the $i$th output, the vector representing the relevant content of the input sequence is obtained as a weighted sum of hidden unit activation vectors:
\begin{equation}
c_i=\sum_{t=1}^{L}\alpha_{i,t} h_t.
\label{eqn:summary}
\end{equation}

The decoder network is an Attention-based Recurrent Sequence Generator (ARSG) \cite{NN_MT@ICLR2015}\cite{NIPS2015_5847} that generates an output label sequence with content vectors $c_{i}$.
The network also has an LSTM decoder network, where the decoder state can be updated in the same way as Equation~(\ref{eqn:decoder-state}).
 
Then, the output label probability is computed as
\begin{equation}
P(y|s_{i-1},c_i)=\text{softmax}\left(W_s^{({\lambda}_D)}s_{i-1} + W_c^{({\lambda}_D)} c_i + b_s^{({\lambda}_D)}\right),
\end{equation}
and word $y_i$ is generated according to
\begin{equation}
y_i=\mathop{\text{argmax}}_{y \in V} P(y|s_{i-1},c_i).
\end{equation}
In contrast to Equations~(\ref{eqn:output-dist}) and (\ref{eqn:output-label}) of the basic encoder-decoder, the probability distribution is conditioned on the content vector $c_i$, which emphasizes specific features that are most relant to predicting each subsequent word.
One more feed-forward layer can be inserted before the softmax layer. In this case, the probabilities are computed as follows:
\begin{equation}
g_i = \tanh\left(W_s^{({\lambda}_D)}s_{i-1} + W_c^{({\lambda}_D)} c_i + b_s^{({\lambda}_D)}\right),
\label{eqn:before-output}
\end{equation}
and
\begin{equation}
P(y|s_{i-1},c_i)=\text{softmax}(W_g^{({\lambda}_D)} g_i + b_g^{({\lambda}_D)}).
\end{equation}

The attention weights are computed in the same manner as in~\cite{NN_MT@ICLR2015}:
\begin{equation}
\alpha_{i,t}=\frac{\exp(e_{i,t})}{\sum_{\tau=1}^{L} \exp(e_{i,\tau})}
\label{eqn:alpha}
\end{equation}
and
\begin{equation}
e_{i,t}=w^{\intercal}_A \tanh(W_A s_{i-1} + V_A h_t + b_A),
\label{eqn:score}
\end{equation}
where $W_A$ and $V_A$ are matrices, $w_A$ and $b_A$ are vectors,
and $e_{i,t}$ is a scalar.

\section{Attention-based multimodal fusion}
This section proposes an attention model to handle fusion of multiple modalities, where each modality has its own sequence of feature vectors.
For video description, multimodal inputs such as image features, motion features, and audio features are available. Furthermore, combination of multiple features from different feature extraction methods are often effective to improve the description accuracy.

In~\cite{attention4spatial@CVPR2016}, content vectors from VGGNet (image features) and C3D (spatiotemporal motion features) are combined into one vector, which is used to predict the next word. This is performed in the fusion layer,
in which the following activation vector is computed instead of Eq. (\ref{eqn:before-output}),
\begin{equation}
g_i = \tanh\left(W_s^{({\lambda}_D)}s_{i-1} + d_i + b_s^{({\lambda}_D)}\right),
\end{equation}
where
\begin{equation}
d_i=W_{c1}^{({\lambda}_D)} c_{1,i} + W_{c2}^{({\lambda}_D)} c_{2,i},
\end{equation}
and $c_{1,i}$ and $c_{2,i}$ are two different content vectors obtained using different feature extractors and/or different input modalities.

\begin{figure}[t]
	\centering
	\centerline{\includegraphics[width=8cm]{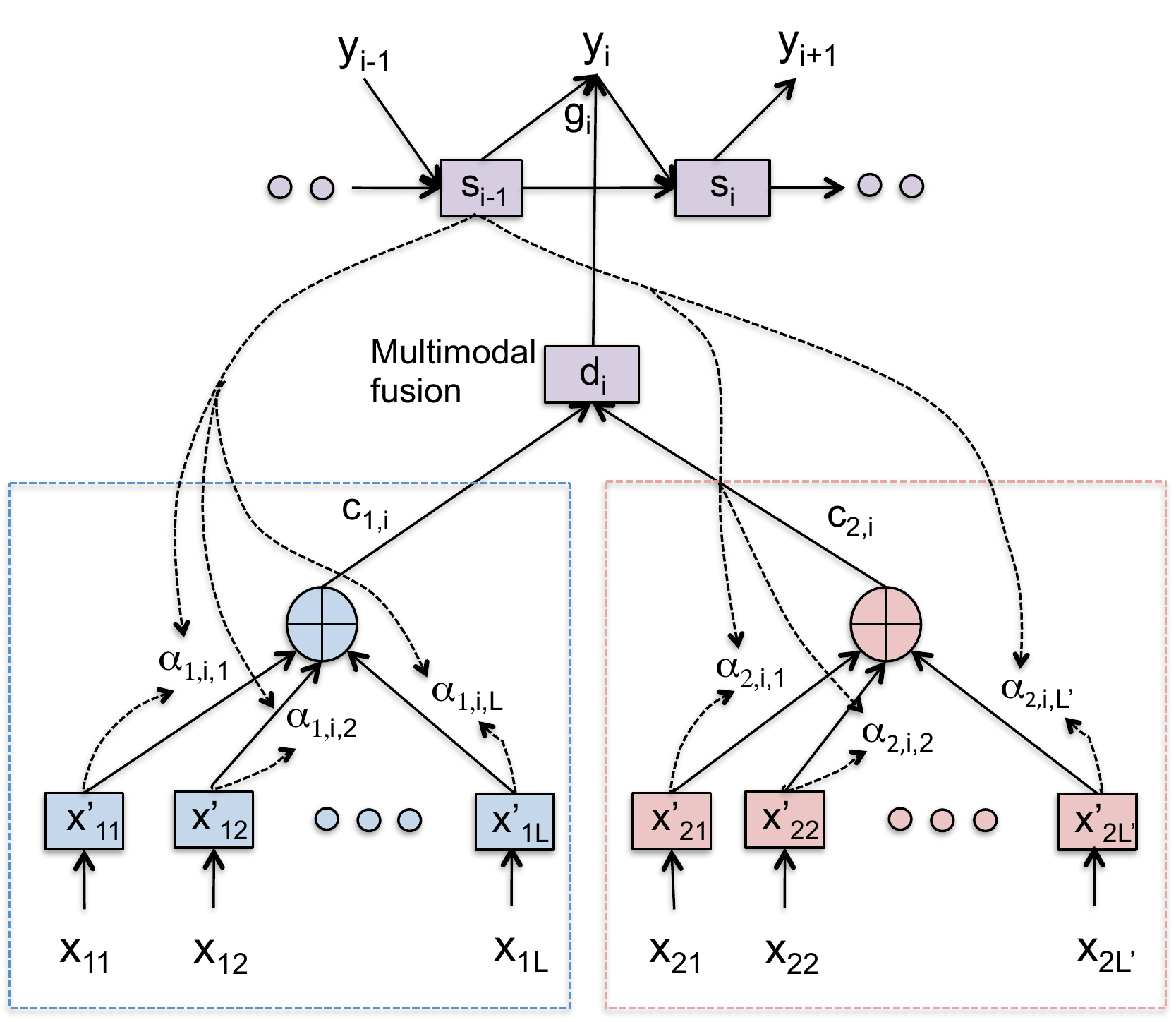}}
	\caption{Simple feature fusion.}
	\label{fig:feature-fusion}
\end{figure}

Figure \ref{fig:feature-fusion} shows the simple feature fusion approach, in which
content vectors are obtained with attention weights for individual input sequences $x_{11},\dots,x_{1L}$ and $x_{21},\dots,x_{2L'}$, respectively.
However, these content vectors are combined with weight matrices $W_{c1}$ and $W_c{2}$, which are commonly used in the sentence generation step. Consequently, the content vectors from each feature type (or one modality) are always fused using the same weights, independent of the decoder state. This architecture lacks the ability to exploit multiple types of features effectively, because it does not allow the relative weights of each feature type (of each modality) to change based on the context.

This paper extends the attention mechanism to multimodal fusion.
Using this multimodal attention mechanism, based on the current decoder state, the decoder network can selectively attend to specific modalities of input (or specific feature types) to predict the next word.
Let $K$ be the number of modalities, i.e., the number of sequences of input feature vectors.
Our attention-based feature fusion is performed using
\begin{equation}
g_i = \tanh\left(W_s^{({\lambda}_D)}s_{i-1} + \sum_{k=1}^K {\beta}_{k,i} d_{k,i}
+ b_s^{({\lambda}_D)}\right),
\end{equation}
where
\begin{equation}
d_{k,i} = W_{ck}^{({\lambda}_D)} c_{k,i} + b_{ck}^{({\lambda}_D)}.
\end{equation}
The multimodal attention weights ${\beta}_{k,i}$ are obtained in a similar way to the temporal attention mechanism:
\begin{equation}
\beta_{k,i}=\frac{\exp(v_{k,i})}{\sum_{\kappa=1}^{K} \exp(v_{\kappa,i})},
\label{eqn:alpha}
\end{equation}
where
\begin{equation}
v_{k,i}=w^{\intercal}_B \tanh(W_B s_{i-1} + V_{Bk} c_{k,i} + b_{Bk}),
\end{equation}
where $W_B$ and $V_{Bk}$ are matrices, $w_B$ and $b_{Bk}$ are vectors,
and $v_{k,i}$ is a scalar.
\begin{figure}[t]
	\centering
	\centerline{\includegraphics[width=8cm]{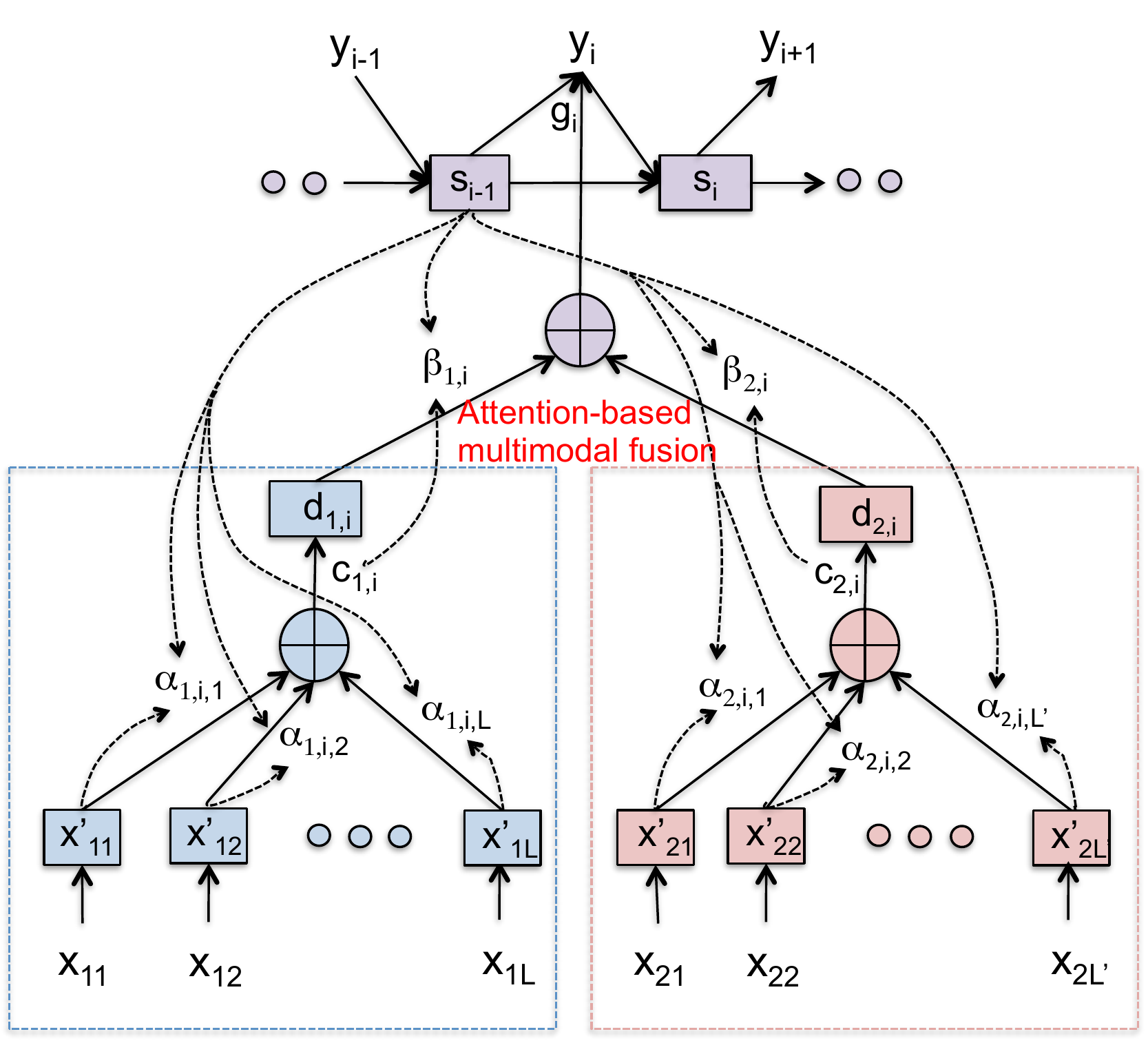}}
	\caption{Our multimodal attention mechanism.}
	\label{fig:feature-attention}
\end{figure}

Figure~\ref{fig:feature-attention} shows the architecture of our sentence generator, including the multimodal attention mechanism. Unlike the simple multimodal fusion method in Figure~\ref{fig:feature-fusion}, in Figure~\ref{fig:feature-attention}, the feature-level attention weights can change according to the decoder state and the content vectors, which enables the decoder network to pay attention to a different set of features and/or modalities when predicting each subsequent word in the description.

\section{Experiments}
\subsection{Dataset}
We evaluated our proposed feature fusion using the Youtube2Text video corpus \cite{guadarrama2013youtube2text}.
This corpus is well suited for training and evaluating automatic video description generation models. The dataset has 1,970 video clips with multiple natural language descriptions.
Each video clip is annotated with multiple parallel sentences provided by different Mechanical Turkers. There are 80,839 sentences in total, with about
41 annotated sentences per clip. Each sentence on average
contains about 8 words. The words contained in all
the sentences constitute a vocabulary of 13,010 unique lexical
entries.
The dataset is open-domain and covers a wide range of topics including sports, animals and music. 
Following [38], we split the dataset into a training set of 1,200 video clips, a validation set of 100 clips, and a test set consisting of the remaining 670 clips.

\setlength\tabcolsep{2pt} 
\begin{table*}[t]
\begin{center}
\caption{Evaluation results on the YouTube2Text test set. The last three rows of the table present previous state-of-the-art methods, which use only temporal attention. The rest of the table shows results from our own implementations. The first three rows of the table use temporal attention but only one modality (one feature type). The next two rows do multimodal fusion of two modalities (image and spatiotemporal) using either Simple Multimodal fusion (see Figure~\ref{fig:feature-fusion}) or our proposed Multimodal Attention mechanism (see Figure~\ref{fig:feature-attention}). The next two rows also perform multimodal fusion, this time of three modalities (image, spatiotemporal, and audio features). In each column, the scores of the top two methods are shown in boldface.}
\label{table:result}
\small
\vspace{4pt}
\begin{tabular}{c||p{1.9cm}|c|c|c||c|c|c|c||c|c}
\hline
	    & & \multicolumn{3}{|c||}{Modalities (feature types)} & \multicolumn{6}{|c}{Evaluation metric} \\
	    \cline{2-11}
	 Fusion method & Attention & Image & Spatiotemporal & Audio & BLEU1 & BLEU2 & BLEU3 & BLEU4 & METEOR & CIDEr \\
\hline
Unimodal ( RMSprop ) & Temporal & GoogLeNet &     &      & 0.766  & 0.643  & 0.547  & 0.440 & 0.295 &  0.568 \\
Unimodal ( RMSprop ) & Temporal & VGGNet    &     &      & 0.800   & 0.677  & 0.574  & 0.464 & 0.309 &  0.654 \\
Unimodal ( RMSprop ) & Temporal &           & C3D &      &  0.785  & 0.664 & 0.569 & 0.464 & 0.304 & 0.578 \\

\hline
\parbox{3cm}{\centering Simple Multimodal \\ ( RMSprop )}  & Temporal & VGGNet & C3D &     &  {\bf 0.824} &  0.708 &  0.606 & 0.498 & 0.322 & 0.665       \\[2pt]
\parbox{3cm}{\centering {\bf Multimodal Attention} \\ ( AdaDelta )} & \parbox{1.9cm}{\bf Temporal \&\\Multimodal}& VGGNet & C3D &    &  0.801 & 0.691 & 0.601 & 0.507 & 0.318 & {\bf 0.699} \\[8pt]
\hline
\parbox{3cm}{\centering Simple Multimodal \\ ( RMSprop )}  & Temporal & VGGNet & C3D & MFCC  &  0.819  & {\bf 0.709} & {\bf 0.614} & 0.510 & 0.321 & 0.679 \\[2pt]
\parbox{3cm}{\centering {\bf Multimodal Attention} \\ ( AdaDelta )} & \parbox{1.9cm}{\bf Temporal \&\\Multimodal} & VGGNet & C3D & MFCC & 0.795 & 0.691 & 0.608 & {\bf 0.517} & 0.317  &  0.695 \\[8pt]
\hline
TA \cite{vd2montreal@ICCV2015}   	& Temporal  &  GoogLeNet    & 3D CNN   &                & 0.800   & 0.647  & 0.526 & 0.419 & 0.296 & 0.517 \\
LSTM-E \cite{LSTM-E} &  & VGGNet & C3D &     & 0.788   & 0.660  & 0.554 & 0.453 & 0.310 & -  \\
h-RNN \cite{attention4spatial@CVPR2016} ( RMSprop )	& Temporal & VGGNet & C3D &     &  0.815 & 0.704 &  0.604 &  0.499 & {\bf 0.326} & 0.658 \\
\hline
\end{tabular}
\end{center}
\end{table*}

\subsection{Video Preprocessing}
The image data are extracted from each video clip, which consist of 24 frames per second, and rescaled to 224$\times$224 pixel images.
For extracting image features, a pretrained GoogLeNet \cite{GoogleNet} CNN is used to extract fixed-length representation with the help of the popular implementation in Caffe \cite{jia2014caffe}. Features are extracted from the
hidden layer pool5/7x7\_s1. We select one frame out of every 16 frames from each video clip and feed them into the CNN to obtain 1024-dimensional frame-wise feature vectors.

We also use a VGGNet~\cite{VggNet} that was pretrained on the ImageNet dataset \cite{NIPS2012_4824}. The hidden activation vectors of fully connected layer fc7 are used for the image features, which produces a sequence of 4096-dimensional feature vectors.
Furthermore, to model motion and short-term spatiotemporal activity, we use the pretrained
C3D~\cite{C3D} (which was trained on the Sports-1M dataset~\cite{karpathy2014large}). The C3D network reads sequential frames in the video and outputs a fixed-length feature vector every 16 frames. We extracted activation vectors from fully-connected layer fc6-1, which has 4096-dimensional features.

\subsection{Audio Processing}
Unlike previous methods that use the YouTube2Text dataset \cite{vd2montreal@ICCV2015,LSTM-E,attention4spatial@CVPR2016}, we also incorporate audio features, to use in our attention-based feature fusion method. Since YouTube2Text corpus does not contain audio track, we extracted the audio data via the original video URLs. Although a subset of the videos were no longer available on YouTube, we were able to collect the audio data for 1,649 video clips, which covers 84\% of the corpus.
The 44 kHz-sampled audio data are down-sampled to 16 kHz, and Mel-Frequency Cepstral Coefficients (MFCCs) are extracted from each 50 ms time window with 25 ms shift. The sequence of 13-dimensional MFCC features are then concatenated into one vector from every group of 20 consecutive frames, which results in a sequence of 260-dimensional vectors. The MFCC features are normalized so that the mean and variance vectors are 0 and 1 in the training set. The validation and test sets are also adjusted with the original mean and variance vectors of the training set.
Unlike with the image features, we apply a BLSTM encoder network for MFCC features, which is trained jointly with the decoder network.
If audio data are missing for a video clip, then we feed in a sequence of dummy MFCC features, which is simply a sequence of zero vectors.

\subsection{Experimental Setup}
The caption generation model, i.e. the decoder network, is trained to minimize the cross entropy criterion using the training set.
Image features are fed to the decoder network through one projection layer of 512 units, while audio features, i.e. MFCCs, are fed to the BLSTM encoder followed by the decoder network.
The encoder network has one projection layer of 512 units and bidirectional LSTM layers of 512 cells.
The decoder network has one LSTM layer with 512 cells.
Each word is embedded to a 256-dimensional vector when it is fed to the LSTM layer.
We compared the AdaDelta optimizer \cite{DBLP:journals/corr/abs-1212-5701} and RMSprop \cite{Tieleman2012} to update the parameters, which is widely used for optimizing attention models.
The LSTM and attention models were implemented using Chainer~\cite{tokui2015chainer}.

The similarity between ground truth and automatic video description results are evaluated using machine-translation-motivated metrics: BLEU~\cite{bleu}, METEOR~\cite{meteor}, and 
the newly proposed metric for image description, CIDEr \cite{CIDEr}. 
We used the publicly available evaluation script prepared for image captioning challenge~\cite{MSCOCO-tool}.
Each video in YouTube2Text has multiple ``ground-truth'' descriptions, but {\em some ``ground-truth'' answers are incorrect.} Since BLEU and METEOR scores for a video do not consider frequency of words in the ground truth, they can be strongly affected by one incorrect ground-truth description. 
METEOR is even more susceptible since it also accepts paraphrases of incorrect ground-truth words. 
In contrast, CIDEr is a voting-based metric that is robust to errors in ground-truth.

\subsection{Results and Discussion}
Table \ref{table:result} shows the evaluation results on the Youtube2text data set.
We compared the performance for our multimodal attention model (Multimodal Attention) which integrated temporal and multimodal attention mechanisms with a simple additive multimodal fusion (Simple Multimodal), unimodal models with temporal attention (Unimodal), and baseline systems that used temporal attention.

The Simple Multimodal model performed better than the  Unimodal models.
The proposed Multimodal Attention model consistently outperformed Simple Multimodal. The audio feature helped the performance of the baseline.
Combining the audio features using our modal-attention method reached the best performance of BLUE.
However, the modal-attention method without the audio feature reached the best performance of CIDEr.
The audio feature did not help always.
This is because some YouTube data includes noise such as background music, which is unrelated to the video content. 
We need to analyze the contribution of the audio feature in detail.

In contrast to the existing systems,
our temporal attention system which used only static image features (Unimodal)
outperformed TA using combination of static image and dynamic video features \cite{vd2montreal@ICCV2015}.
Our proposed attention mechanisms outperformed LSTM-E \cite{LSTM-E} which does not use attention mechanisms. 
Our Simple Multimodal system using temporal attention is the same basic structure used by h-RNN as well as the same features extracted from VGGNet \cite{VggNet} and C3D \cite{C3D}.
While h-RNN used L2 regularization and RMSprop, 
we used L2 regularization for all experimental conditions and compared RMSprop and AdaDelta. 
Although RMSprop outperformed for Umimodal and Simple Multimodal,
AdaDelta outperformed for {\bf Multimodal Attentiontion}.


\section{Conclusion}
We proposed a new modality-dependent attention mechanism, which we call multimodal attention, 
for video description based on encoder-decoder sentence generation using recurrent neural networks (RNNs).
In this approach, the attention model selectively attends not just to specific times, 
but to specific modalities of input such as image features, spatiotemporal motion features, and audio features. 
This approach provides a natural way to fuse multimodal information for video description.
We evaluate our method on the Youtube2Text dataset, achieving results that are competitive with current state-of-the-art methods that employ temporal attention models, 
in which the decoder network predicts each word in the description by selectively giving more weight to encoded features 
from specific time frames. 
More importantly, we demonstrate that our model incorporating multimodal attention 
as well as temporal attention outperforms the model that uses temporal attention alone.

{\small
\bibliographystyle{ieee}
\bibliography{egbib}
}

\end{document}